\documentclass{nature}
\usepackage{graphicx}
\usepackage{sectsty}
\usepackage{amssymb} 
\usepackage{amsmath} 
\usepackage{mathtools}
\usepackage{bm}  
\usepackage{cite}
\usepackage[ruled,vlined]{algorithm2e}
\usepackage{hyperref}
\usepackage{siunitx}
\usepackage{doi}
\usepackage{url}
\usepackage{tabularx}
\usepackage{xltabular}
\usepackage{caption}
\usepackage{subcaption}
\usepackage[dvipsnames]{xcolor}
\usepackage[shortlabels]{enumitem}
\usepackage[export]{adjustbox}
\usepackage{multirow}
\usepackage[percent]{overpic}
\usepackage{colortbl}
\usepackage{pgfplots}
\pgfplotsset{compat=newest}

\usepackage{color}

\renewcommand{\vec}[1]{\boldsymbol{\mathbf{#1}}}

\definecolor{matlabcyan}{rgb}{0 1 1}
\definecolor{matlabmagenta}{rgb}{1 0 1}

\title{Real-time Non-line-of-Sight imaging of dynamic scenes}
\author{Ji Hyun Nam$^{1}$, Eric Brandt$^{2}$, Sebastian Bauer$^{3}$, Xiaochun Liu$^{1}$, Eftychios Sifakis$^{2}$, Andreas Velten$^{1,3}$}

\begin{document}

\maketitle

\begin{affiliations}
\item Department of Electrical and Computer Engineering, University of Wisconsin Madison
\item Department of Computer Science, University of Wisconsin Madison
\item Department of Biostatistics and Medical Informatics, University of Wisconsin Madison
\end{affiliations}

\begin{abstract} 
Non-Line-of-Sight (NLOS) imaging\cite{Velten_12} aims at recovering the 3D geometry of objects that are hidden from the direct line of sight. In the past, this method has suffered from the weak available multibounce signal limiting scene size, capture speed, and reconstruction quality. While algorithms capable of reconstructing scenes at several frames per second have been demonstrated,\cite{liu_phasor_2020,Lindell_19_Wave, OToole_18} real time NLOS video has only been demonstrated for retro-reflective objects where the NLOS signal strength is enhanced by 4 orders of magnitude or more.
Furthermore, it has also been noted that the signal to noise ratio of reconstructions in NLOS methods drops quickly with distance and past reconstructions therefore have been limited to small scenes with depths of few meters. Actual models of noise and resolution in the scene have been simplistic, ignoring many of the complexities of the problem.
We show that SPAD (Single-Photon Avalanche Diode) array detectors with a total of just 28 pixels combined with a specifically extended Phasor Field reconstruction algorithm\cite{liu2019virtual,liu_phasor_2020,reza2018physical,reza2019wave} can  reconstruct live real time videos of non-retro-reflective NLOS scenes. We provide an analysis of the Signal-to-Noise-Ratio (SNR) of our reconstructions and show that for our method it is possible to reconstruct the scene such that SNR, motion blur, angular resolution, and depth resolution are all independent of scene size suggesting that reconstruction of very large scenes may be possible. In the future, the light efficiency for NLOS imaging systems can be improved further by adding more pixels to the sensor array.
\end{abstract}

\section*{}
The ability to image around corners has attracted significant interest in the research community after its first analysis~\cite{kirmani2009looking} and  demonstration~\cite{Velten_12}. It offers applications in many spaces such as autonomous vehicle navigation, collision avoidance, disaster response, infrastructure inspection, military and law enforcement operations, mining, and construction. While other methods have been investigated as well \cite{Heide_14,bouman2017turning,smith2018tracking,saunders2019computational}, robust reconstructions of near room sized scenes have only been demonstrated using methods that rely on photon Time of Flight (ToF). In ToF methods a short laser pulse is focused on a point $\vec x_\mathrm{p}$ on the diffuse relay surface from where the light scatters in many directions. Some photons travel to objects hidden from the direct line of sight and get reflected back to the relay surface. A fast photo detector images a patch $\vec x_\mathrm{c}$ on the relay surface recording the time of arrival of photons reflected off $\vec{x}_\mathrm{c}$. Scanning the relay surface point $\vec{x}_\mathrm{p}$ with the laser provides sufficient information for 3D reconstruction of the hidden scene, with some limitations \cite{LiuX_19}. Scene reconstruction is possible using a variety of methods \cite{La_18,Xin_19,ahn2019convolutional,Tsai_19,Chen_20}; see \cite{faccio2020non} for a thorough review. 

Existing NLOS methods suffer from a very low light efficiency as only a very small fraction of the illumination light is eventually detected and can contribute to the reconstruction.

While fast reconstruction algorithms have been demonstrated~\cite{Lindell_19_Wave,OToole_18,liu_phasor_2020}, the low light levels in the captured signal, have so far resulted in capture times of minutes~\cite{Lindell_19_Wave,OToole_18} or tens of seconds~\cite{liu2019virtual} for general scenes which is not acceptable in most applications. The only real time reconstructions to date~\cite{Lindell_19_Wave,OToole_18} reconstruct retro-reflecting surfaces, that for the specific scenes  and using a specialized confocal scanning capture technique provide signals at least 10,000 times higher than diffuse surfaces in the demonstrated geometries and therefore are not indicative of NLOS performance in many real scenes. 

\begin{figure}[hb!]
\centering
\includegraphics[width=\linewidth]{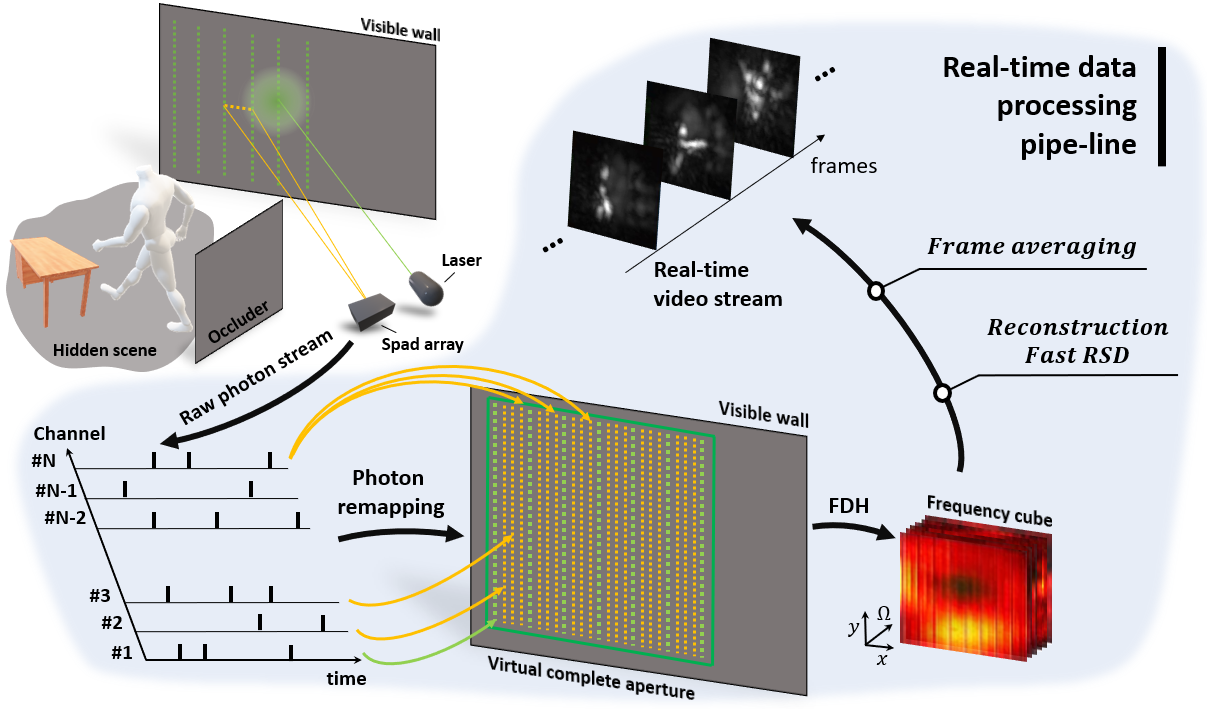}
\caption{Real-time NLOS virtual image processing pipeline. The imaging system sends the virtual PF signal to the visible wall and captures the signal returning from the hidden scene back to the wall. The massive raw photon stream is recorded by the SPAD array. Raw photons from all channels are virtually remapped into a complete aperture. Then the remapped data is transformed into the frequency domain and propagated by the fast RSD. Lastly, temporal frame averaging yields constant SNR throughout the entire reconstructed volume, and the result is displayed.}
\label{fig:main_plot}
\end{figure}

Reconstructed scene sizes have been limited to few meters in diameter and it has been speculated that objects further from the relay surface would be much harder to reconstruct based on the strong distance dependence of the returned signal strength~\cite{Lindell_19_Wave,faccio2020non}. In this work we demonstrate how to overcome these spatial and temporal constraints and create a real-time NLOS video reconstruction method that extends to larger scene distances.

Let us examine the behavior of the NLOS signal as a function of object distance in the scene. Past work states that the signal from a small, fixed-size patch in the hidden scene collected from an individual pair of co-located laser and detector positions $\vec x_\mathrm{p}, \vec x_\mathrm{c}$  near the center of the relay surface falls off as $1/r^4$ for the shortest distance $r$ between relay surface and object. However, for the cumulative signal from a complete NLOS measurement comprising a set of $\vec x_\mathrm{p}$ and $\vec x_\mathrm{c}$ this is only true for very large $r$ and does not hold at close distances that apply for most reconstructions. At such close distances, the falloff is smaller than $1/r^4$. A detailed mathematical analysis of the falloff is provided in the Supplementary Materials. Furthermore, the reduction in resolution at large distances makes considering a patch of fixed size misleading as the patch simply drops below the resolution limit. Finally, existing investigations consider only the drop in collected signal and ignore the change of noise as a function of $r$. In most conventional optical imaging systems, noise is considered for a fixed angular resolution where the scene area corresponding to an image pixel increases with distance along with the imaging system resolution. For example, in a conventional camera, noise is added to the image at the camera sensor, after image reconstruction has been been performed using a lens. By contrast, in an NLOS imaging system, Poisson and sensor noise occur in the measurement \textit{before} application of the image reconstruction inverse operator. This noise is then propagated through the reconstruction operator which essentially mimics the operation of the imaging lens. As a consequence the noise in a NLOS reconstruction is different from the noise in a line of sight image. In particular, it depends on distance $r$. 

We show here that appropriately designed SPAD array detectors and reconstruction methods can be used to overcome the deficits in SNR of NLOS imaging and enable real time, low latency NLOS video with depth independent SNR and motion blur. This results in a constant observable motion speed, angular and depth resolution,  and a constant SNR throughout the reconstructed scene.

Even though a NLOS imaging experiment illuminates only a single point on the relay surface at a time, the returning signal is distributed over the entire relay surface. Past NLOS methods have relied on single pixel SPAD detectors or SPAD arrays with very small pixels, low time resolution and no gate~\cite{gariepy2016detection} that collected only light from a small fraction of the relay surface. By utilizing an appropriately designed SPAD array, the amount of signal collected can be increased by increasing the relay surface area from which light is collected.

\begin{figure}[t]
\centering
\includegraphics[width=\linewidth]{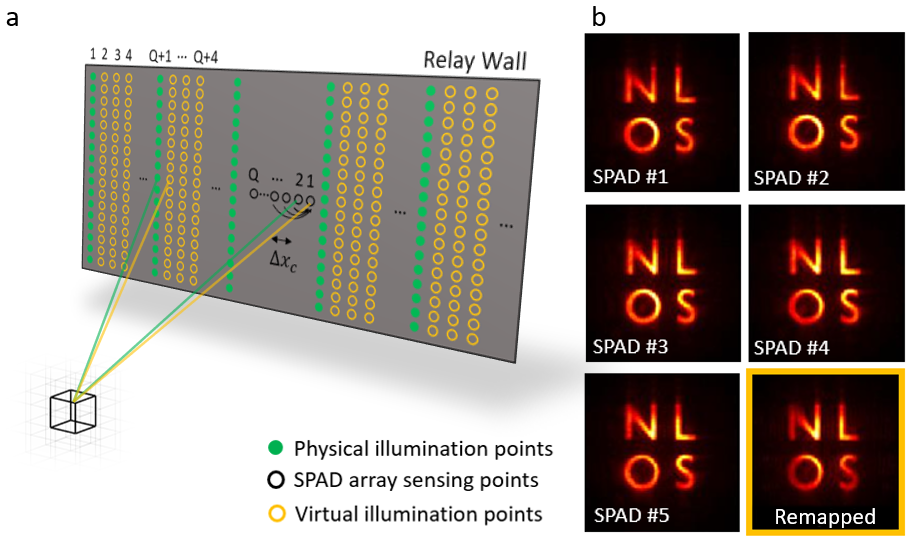}
\caption{Virtual aperture remapping operation. a) Green points are the physically scanned illumination positions and black points are the SPAD array's sensing positions. Data acquired by the SPAD array is remapped into a full virtual aperture. b) 5 reconstructions of letters from 5 separate SPAD pixels and 1 reconstruction using virtually remapped data from 5 SPAD pixels.}
\label{fig:scanpattern_new}
\end{figure}

To improve the light efficiency of our setup we use two 16 by 1 pixel fast gated SPAD arrays~\cite{Renna20} that image light from a line of patches on the relay surface. In principle one could scan the relay surface in a continuous grid and use an existing phasor field~\cite{liu2019virtual} or filtered back-projection~\cite{Velten_12} method to perform the reconstruction on the data acquired by SPAD arrays. However, the phasor field and back-projection algorithms are not able to reconstruct the scene fast enough for creating real time videos. By contrast, we develop a Phasor Field (PF) reconstruction algorithm that allows for sparse scanning of the relay surface and can make use of the fast convolutional phasor reconstruction method based on Rayleigh Sommerfeld Diffraction (RSD) introduced by Liu et al.\cite{liu_phasor_2020}. Our proposed real-time NLOS video processing pipeline, which includes a fast capture and reconstruction scheme, is illustrated in Figure~\ref{fig:main_plot}.

To begin,  we introduce a sparse illumination scanning pattern with a remapping operation based on SPAD array sensing points to create a virtual complete continuous illumination grid with a single sensing point. This approach has several benefits. First, the sparse illumination pattern reduces the physical scan time allowing to achieve high capturing frame rate, which is essential when it comes to dynamic scenes. Second, the RSD method\cite{liu_phasor_2020} assumes a continuous illumination grid $\vec{x}_\mathrm{p}=(x_\mathrm{p},y_\mathrm{p},0)$ and a single sensing point $\vec{x}_{\mathrm{c}}=(x_{\mathrm{c}},y_{\mathrm{c}},0)$. Therefore transient data collected from multiple sensing points  $\vec{x}_{\mathrm{c},q}=(x_{\mathrm{c},q},y_{\mathrm{c},q},0)$, $q \in [1,\hdots,Q]$ using a SPAD array cannot be  used directly. The remapping operation addresses this issue by virtually remapping the sparsely scanned grid $\vec{\bar{x}}_\mathrm{p}$ and multiple sensing points $\vec{x}_{\mathrm{c},q}=(x_{\mathrm{c},q},y_{\mathrm{c},q},0)$ into a  complete illumination grid $\vec{x}_\mathrm{p}=(x_\mathrm{p},y_\mathrm{p},0)$ and single sensing point $\vec{x}_{\mathrm{c},1}=(x_{\mathrm{c},1},y_{\mathrm{c},1},0)$. The remapping operation exploits a spatial relationship between $\vec{\bar{x}}_\mathrm{p}$ and the $\vec{x}_{\mathrm{c},q}$, and approximates the missing illumination positions which is illustrated in Fig.~\ref{fig:scanpattern_new}a. Mathematically, given a set of sparse laser positions $\vec{\bar{x}}_\mathrm{p}$ and a set of SPAD positions $\vec{x}_{\mathrm{c},q}$, one seeks to obtain the measurement at a laser grid shifted by small amounts $\Delta \vec{x}_{\mathrm{c},q}=(\Delta x_{\mathrm{c},q},0,0)$. This can be done by virtually remapping the captured data to shifted locations and approximating the required time responses as
\begin{equation}\label{eq:approx_rearrangement}
H((\vec{\bar{x}}_\mathrm{p}+\Delta \vec{x}_{\mathrm{c},q})\rightarrow \vec{x}_{\mathrm{c},1}, t) \approx H(\vec{\bar{x}}_\mathrm{p}\rightarrow (\vec{x}_{\mathrm{c},1}-\Delta \vec{x}_{\mathrm{c},q}), t), \quad q \in [1,\hdots,Q].
\end{equation}
This holds as long as the length of the spatial shift $\Delta x_\mathrm{c,q}$ is small with respect to the distance between the laser position on the relay wall and the object location, and the object location and the SPAD position on the relay wall, respectively. The transient data acquired by the leftmost SPAD pixels are projected to the rightmost SPAD pixel location, which virtually creates a full illumination scan grid. The virtually remapped transient data $H((\vec{\bar{x}}_\mathrm{p}+\Delta x_{\mathrm{c},q})\rightarrow \vec{x}_{\mathrm{c},1}, t)$ can fully utilize the fast RSD method allowing for real-time reconstruction:
\begin{align}
\label{eq:rsdreconstruction}
    \begin{split}
    \mathcal{I}(x_\mathrm{v},y_\mathrm{v},z_\mathrm{v}) &=
    \Phi_F(\mathcal{P_F}(\vec{x}_\mathrm{p},t) * H((\vec{\bar{x}}_\mathrm{p}+\Delta \vec{x}_{\mathrm{c},q})\rightarrow \vec{x}_{\mathrm{c},1}, t)),
    \end{split}
\end{align}
where $\Phi_F(\cdot)$ is the image formation function and $\mathcal{P_F}$ is the Phasor Field illumination function Fourier-transformed with respect to time. Figure~\ref{fig:scanpattern_new}b shows 5 different reconstructions from 5 individual SPAD pixels using the full illumination pattern, and one reconstruction using the virtually remapped illumination pattern from the sparse illumination pattern with 5 SPAD array pixels. The reconstruction result from virtually remapped data has comparable quality to single pixel reconstructions while allowing for significantly faster and more efficient data acquisition.

By consecutively scanning the relay wall rapidly with the sparse illumination grid, we get a sequence of reconstructions $\mathcal{V}(\tau)=\mathcal{I}(x_\mathrm{v},y_\mathrm{v},z_\mathrm{v}; \tau)$, $\tau=[1,2,..]$ with exposure time of $\Delta\tau$ seconds for each frame $\tau$. The frame rate of $\mathcal{V}(\tau)$ is $1/\Delta\tau$.

Let us now examine the application of our capture method and noise model to the reconstruction of large dynamic scenes. It has been shown that motion of the relay surface and thus the virtual camera and physical imaging system can be compensated~\cite{la2020non}. We therefore assume that only motion of objects within the hidden scene can prevent accurate reconstruction. Furthermore we will assume a maximum object velocity we seek to be able to image. 

\begin{figure}[hbt!]
\centering
\includegraphics[width=\linewidth]{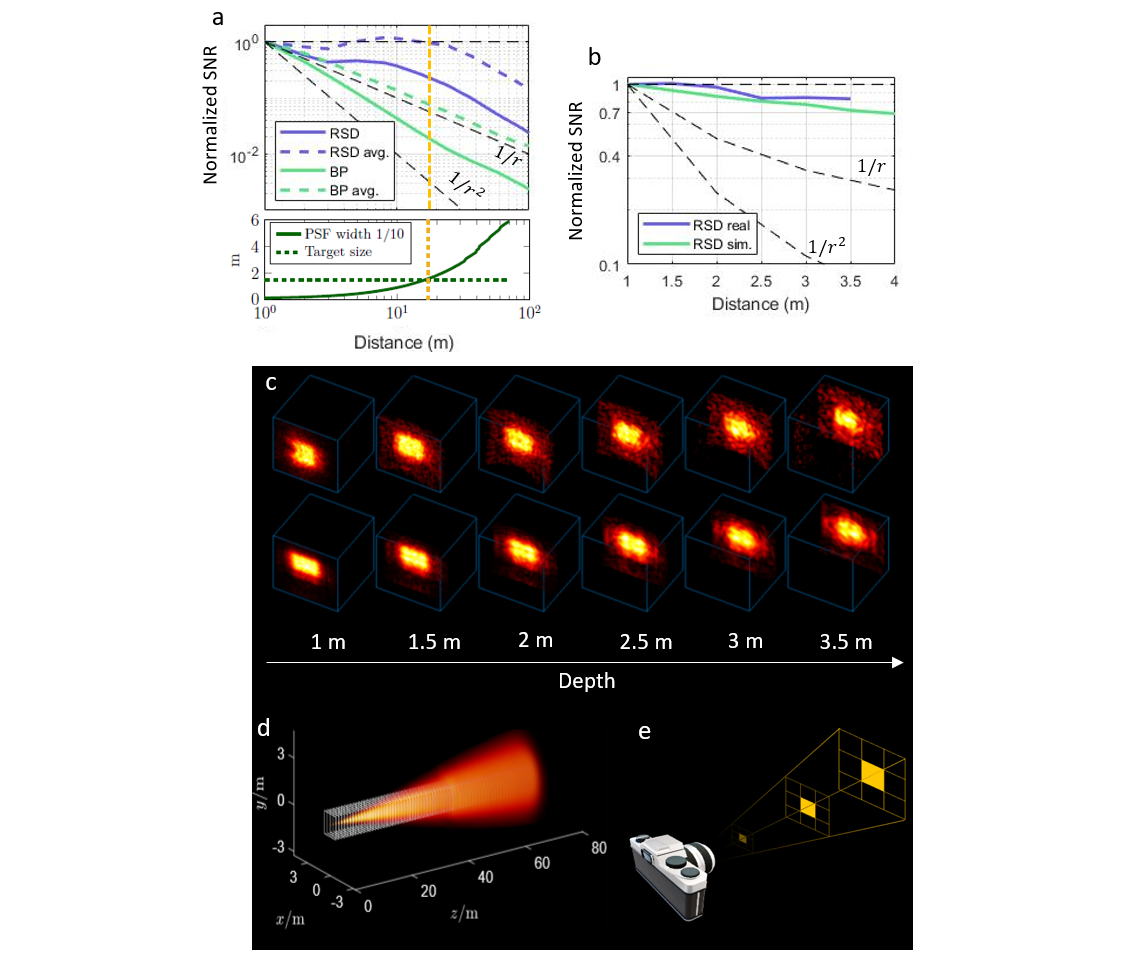}
\caption{NLOS SNR. a: simulated SNR for a square target of \SI{1.5}{m} width, no ambient light. b: real data as well as simulated results with ambient light level seen by the current system with lights on. c: reconstructions from real (top) and simulated data (bottom) at different depth. All SNR curves have been normalized to a value of 1 at \SI{1}{m} to only show the falloff with distance; the absolute values are parameter dependent. d: central depth slice of RSD PSF at different depths. The target size is shown as white squares. e: pixel size in conventional photography.}
\label{fig:SNR_main}
\end{figure}

The spatial resolution of an NLOS image can be described by a three dimensional Point Spread Function (PSF) $\mathrm{PSF}(x, y, z)$. The widths $\Delta x, \Delta y,\Delta z$ of the PSF along each dimension indicate the resolution of the reconstruction in the different dimensions. It has been shown~\cite{OToole_18} that the achievable NLOS imaging resolution grows proportionally with distance. The Phasor Field reconstruction PSF matches this increase  approximately following the Rayleigh criterion resulting in a constant angular resolution. This is shown using simulated PSFs in Figure~\ref{fig:SNR_main}d. This means that the Phasor Field reconstructions are automatically filtered with a depth dependent low-pass filter that matches the achievable image resolution. This improves the SNR at larger distances compared to other methods.

It is beneficial to express the $\mathrm{PSF}$ in spherical coordinates: $\mathrm{PSF}(\phi, \theta, r)$ with widths $\Delta \phi, \Delta \theta, \Delta r$ which are largely independent of location in the scene. To incorporate moving objects, we assume that motion blur will only be noticeable when the object in question moves by about the size of the PSF in a given frame. If the motion is smaller the motion blur kernel will be negligible to the PSF and will not significantly affect the reconstruction. This means that for a given maximum object velocity $v$, the required exposure time is proportional to $\Delta \phi/v_{\phi}, \Delta \theta/v_{\theta}, \Delta r/v_{r}$ and increases linearly with distance $r$. Consequently, we can safely average scene voxels at distances far from the relay wall over longer times without expecting visible motion blur. Objects far away from the camera appear to be moving slower. This is not unique to NLOS reconstructions, but is the feature of any video. In NLOS reconstructions, however, we have the option to choose a depth dependent frame averaging in the reconstruction to take advantage of this effect. We create a reconstruction in which the frame rate is depth dependent. The details of our depth dependent frame averaging method are described in the Methods section.

We use simulated and real captured data to evaluate the SNR of a Phasor Field reconstruction that includes: a) the depth dependent intensity; b) depth dependent noise that is the result of passing poisson and ambient light noise through the reconstruction operator (see supplement for the details of the noise model); c) the depth dependent frame averaging along with the depth dependent band filtering inherent to the Phasor Field algorithm  (see Figure \ref{fig:SNR_main}). In all cases, we placed a planar diffuse white patch (\SI{1.5}{m} x \SI{1.5}{m}) in the hidden scene at different depths and collected 100 repeated measurements in the real experiment (depths from \SI{1}{m} to \SI{3.5}{m}) and \SI{20000}{} measurements in the simulated case (depths up to \SI{500}{m}). Each dataset was processed and reconstructed individually; Fig.~\ref{fig:SNR_main}c shows exemplary  reconstructions. All individual reconstruction results are used to evaluate the SNR defined as the mean over the standard deviation at different depths by calculating the sample mean and sample standard deviation over multiple noisy reconstructions. Figures \ref{fig:SNR_main}a,b show the SNR of the reconstruction at different depths for simulated and real data. For more detailed plots please see the Supplement.

As we can see in Figure~\ref{fig:SNR_main}a, the SNR of backprojection (BP) decreases rapidly for large distances, whereas SNR of the phasor field RSD reconstruction decreases slower compared to BP because of the inherent spatial averaging which compensates the SNR loss. Lastly, after applying the optimal depth dependent frame averaging introduced above we find that the SNR of RSD stays constant approximately up to the distance where the target becomes smaller than the resolution limit of our imaging system.

\begin{figure}[ht!]
\centering
\includegraphics[width=\linewidth]{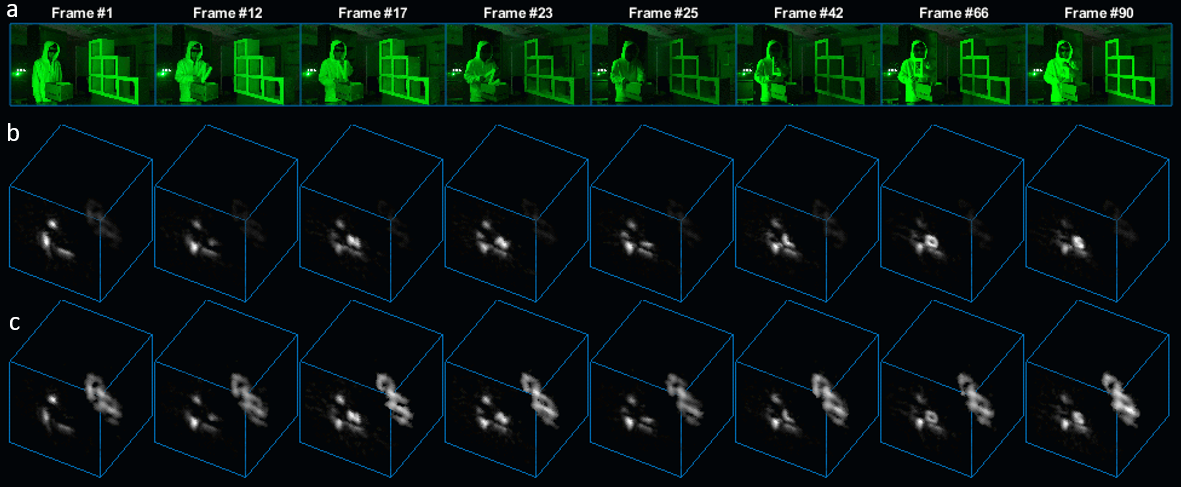}
\caption{Real-time video: "NLOS letter box". 8 sample frames from a 20 second real-time video. The person takes four letters (N, L, O, S) out of a box. The first row shows ground truth images of the hidden scene. The second row shows reconstructed virtual frames using RSD. The third row shows reconstructed virtual frames after applying depth dependent frame averaging.}
\label{fig:video_1}
\end{figure}

Finally, we are ready to present NLOS videos that are both acquired and reconstructed in real-time using the proposed method.
To demonstrate real world capabilities we first implemented the optimized pipeline of our proposed method (Figure ~\ref{fig:pipeline}), and second, a hardware system with two SPAD arrays and laser has been designed. See the Methods section for implementation details and hardware specifications. Figure~\ref{fig:video_1} shows several frames of a dynamically moving complex NLOS scene captured and reconstructed in real-time using our system. Our imaging system scans a sparse relay wall scan pattern with 190x22 sampling points. Scanning rate is 5 frames per second (fps), hence, the exposure time per frame is \SI{0.2}{s}. The remapped complete virtual aperture has size \SI{1.9}{m} x \SI{1.9}{m} with 190x190 virtual sampling points. The Phasor Field virtual wavelength is set to \SI{8}{cm}. Our NLOS imaging system captures data, reconstructs the dynamic hidden scene and displays the result to the user in real-time. The computational pipeline is acquisition-bound and easily supports a throughput of 5 fps with a latency of 1 second. The target SNR was set to be equal to the SNR at distance $z_{0}=\SI{1}{m}$. Hence, at largest depth $z_\mathrm{max}=\SI{3}{m}$ of our reconstruction we average 3 frames, which is factored into the computational latency mentioned above. The maximum motion velocity is given by the \SI{200}{ms} exposure time and the \SI{8}{cm} spatial resolution and is about 0.4 meters per second  which is sufficient to capture normal human movements.  Figure~\ref{fig:video_1}a shows the ground truth of the dynamic hidden scene, and reconstructed NLOS frames are shown in Fig.~\ref{fig:video_1}b. Figure~\ref{fig:video_1}c shows the result after the depth dependent frame averaging has been applied. More video results are depicted in the Supplement. Additionally, readers are encouraged to view the full real-time NLOS videos in the supplementary video file. 

We have shown that the capabilities of NLOS imaging systems can be substantially increased with the combinations of purposefully designed SPAD arrays and array specific fast reconstruction algorithms. Using a total of only 28 pixels we reconstruct real time NLOS videos of non-retro-reflective objects for the first time. We also show that scene size and object distance do not represent an insurmountable problem for NLOS imaging. While we believe that this is a major step forward in the demonstration of the capability of NLOS imaging and towards the actual deployment of NLOS imaging systems in real-world applications such as robot navigation, disaster response and many others, there are still opportunities for future work. CMOS SPAD array technology allows for the fabrication of kilopixel and recently even megapixel arrays at low cost. We expect that future NLOS imaging systems will further improve capabilities by adding more pixels to improve SNR, speed, and stand-off distance, as well as increase the relay wall size to improve reconstruction resolution.

\newpage

\section*{Methods}

\textbf{Depth-dependent frame averaging} 

To compensate for the SNR decrease (Figure ~\ref{fig:SNR_main}a), we can apply linear depth dependent frame averaging. Note that one can choose the target SNR level arbitrarily. Without loss of generality, we choose this level to be the SNR at the  distance $z_{0}$ that is closest to the relay wall. For the reconstruction slice $\mathcal{I}(x_\mathrm{v},y_\mathrm{v},z_\mathrm{v}; \tau)$ at $z_\mathrm{v}=z_\mathrm{0}$ no averaging is needed. For all consecutive depths $z_\mathrm{i}, i=[1,2,3,\hdots]$, we take the average of the $N = \mathrm{ceil}(z_\mathrm{i})$ past frames, that is,

\begin{align}
\label{eq:frameaveraging}
    \begin{split}
    \mathcal{I}_\mathrm{avg}(x_\mathrm{v},y_\mathrm{v},z_\mathrm{i};\tau)=\frac{1}{N}\sum_{n=0}^{N-1}  \mathcal{I}(x_\mathrm{v},y_\mathrm{v},z_\mathrm{i};\tau-n)\,.
    \end{split}
\end{align}

For the Phasor Field method, we apply averaging to both the real and imaginary parts first before taking the absolute value squared (coherent summation). As a consequence, the statistical mean value of the averaged reconstruction does not equal the statistical mean of a single reconstruction, see Supplementary Figures 3 and 4. 

In addition to the depth dependent frame averaging we can compensate the intensity decrease by applying depth dependent intensity correction through multiplication by the mean values resulting from the SNR calculations (Supplementary Figures 3 and 4).   

\textbf{Details on the hardware configuration, calibration and acquisition} 

Core components of our imaging system are the ultra-fast laser and the SPAD array. The used laser is an OneFive Katana HP pulsed laser operating at 532 nm with a pulse width of \SI{35}{ps}. The operating laser power is \SI{700}{mW} with a repetition rate of \SI{5}{MHz}. 

The prototype SPAD array has 16 pixels with a temporal resolution of \SI{75}{ps} and a dead time of \SI{200}{ns}. The active gate window duration can be adjusted. We set it to \SI{40}{ns}, which results in a round trip of roughly \SI{6}{m} during which the sensor is not registering any photons and therefore disregards the bright first reflection off the relay wall. In this work we use two 16 by 1 pixel SPAD arrays~\cite{Renna20}, placed horizontally in a row in the imaging system, see Fig.~\ref{fig:hardware}. Both SPAD arrays are focused in the middle of relay wall using a Nikon \SI{50}{mm} F1.2 objective lenses. Each pixel's observation area on the wall is approximately \SI{5}{mm^2}. The width of the total SPAD array focus area on the relay wall is approximately \SI{8}{cm}. One Thorlabs FL532-3 band pass filter at \SI{532}{nm} with FWHM \SI{3}{nm} is placed in front of each SPAD array to reject ambient light of different wavelengths. 

As a photon counting device we use the PicoQuant HydraHarp 400 Time-Correlated Single Photon Counting (TCSPC) unit with eight channels. Here we use the Time-Tagged Time-Resolved (TTTR) mode for the data acquisition at \SI{8}{ps} time resolution. Combined, the effective temporal uncertainty of laser and each SPAD pixel is approximately \SI{85}{ps}. One HydraHarp channel is used for the confocal single pixel SPAD for system calibration. The remaining 7 channels are used by both SPAD arrays; since the laser's repetition rate is \SI{5}{MHz}, each HydraHarp channel has an available time window of \SI{200}{ns} before the next laser pulse. 4 pixels from the SPAD arrays are connected to one TCSPC channel. In order to separate the signals from the 4 SPAD pixels within a single TCSPC channel, we use cables of corresponding length to delay the signals from each SPAD pixel. Thus, the \SI{200}{ns} time window is divided into four sections: [\SI{0} - \SI{40}{ns}], [\SI{40} - \SI{90}{ns}], [\SI{90} - \SI{140}{ns}] and [\SI{140} - \SI{180}{ns}]. In total we utilize $7 \times 4 $ = 28 SPAD pixels.

We scan illumination points on the relay wall using a set of two mirror galvanometers (Thorlabs GVS012). The maximum frequency of this system is \SI{150}{Hz}, meaning that 150 vertical lines per second can be scanned. The laser scans the relay wall continuously in a raster pattern. Our physical laser grid has 190x22 points. Vertical and horizontal spacing is \SI{1}{cm} and \SI{9}{cm} respectively. The full laser grid scanning rate is \SI{5}{fps} and the exposure time per frame is \SI{0.2}{s}, meaning that each laser point is exposed for \SI{480}{microseconds}. See the Supplement for examples of acquired data. 

Our imaging system is located about \SI{2}{m} away from the relay wall. To calibrate the system we use a single pixel SPAD that is aligned with the laser beam path (see Figure~\ref{fig:hardware}). We scan the relay wall (\SI{1.9}{m} x \SI{1.9}{m}) and the single pixel SPAD collects direct light from the relay wall, which yields the distances from the imaging system to the illumination points. Next, we scan a small square region around the SPAD array sensing points and collect the signal with SPAD array. The collected data is used to evaluate the distances from imaging system to the SPAD array sensing points on the relay wall. These distances are used to virtually shift the collected data to the relay wall.

The hidden scene starts at \SI{1}{m} away from the relay wall and goes up to  \SI{3.5}{m} as limited by the time ranges provided by the TCSPC. The scene consists of conventional diffuse objects. The person in the scene is wearing a regular white hooded sweatshirt. Figures~ \ref{fig:extra_1b}, \ref{fig:extra_1c}, \ref{fig:extra_3}, \ref{fig:extra_4}, \ref{fig:extra_5} show examples of different scenes and reconstructions. These figures also contain depth dependent intensity corrected results. NLOS video results of corresponding scenes can be found in the supplementary video file.  

\textbf{Details on the implementation of real-time reconstruction pipeline} 

Our custom software implements a processing pipeline that reads incoming photon data from the hardware, performs the NLOS image reconstruction, and displays resulting 2D images to the screen in real-time, keeping pace with hardware acquisition rates described above. The software is written in C++ and makes use of both CPU multithreading and GPU computation.

\begin{algorithm}[!ht]
\SetAlgoLined
\PrintSemicolon
\SetKwInOut{Input}{input}\SetKwInOut{Output}{output}
\SetKwFunction{ZeroMemory}{ZeroMemory}
\Input{Array, $photons$, of $(gridIdx, photonTime)$ tuples \\ 
       Array, $freqs$, of frequencies}
\Output{Array $fdh[\texttt{NUM\_FREQS}][\texttt{NUM\_GRID\_INDICES}][2]$, Fourier Domain Histogram }
\BlankLine
\ZeroMemory{$fdh$}\;
\ForPar{$f=0$ \KwTo \texttt{NUM\_FREQS}}{
    \ForEach{$(idx, time)$ in $photons$}{
        $fdh[f][idx][0]$ += SinLookupTable[(int)($freqs[f]$ * $time$)]\;
        $fdh[f][idx][1]$ += CosLookupTable[(int)($freqs[f]$ * $time)$]\;
    }
}
\Return $fdh$\;
\caption{Frequency Domain Histogram Binning (un-optimized for clarity)}
\label{alg:FDH}
\end{algorithm}

\begin{figure}[htb]
    \centering
    \includegraphics[width=0.8\linewidth]{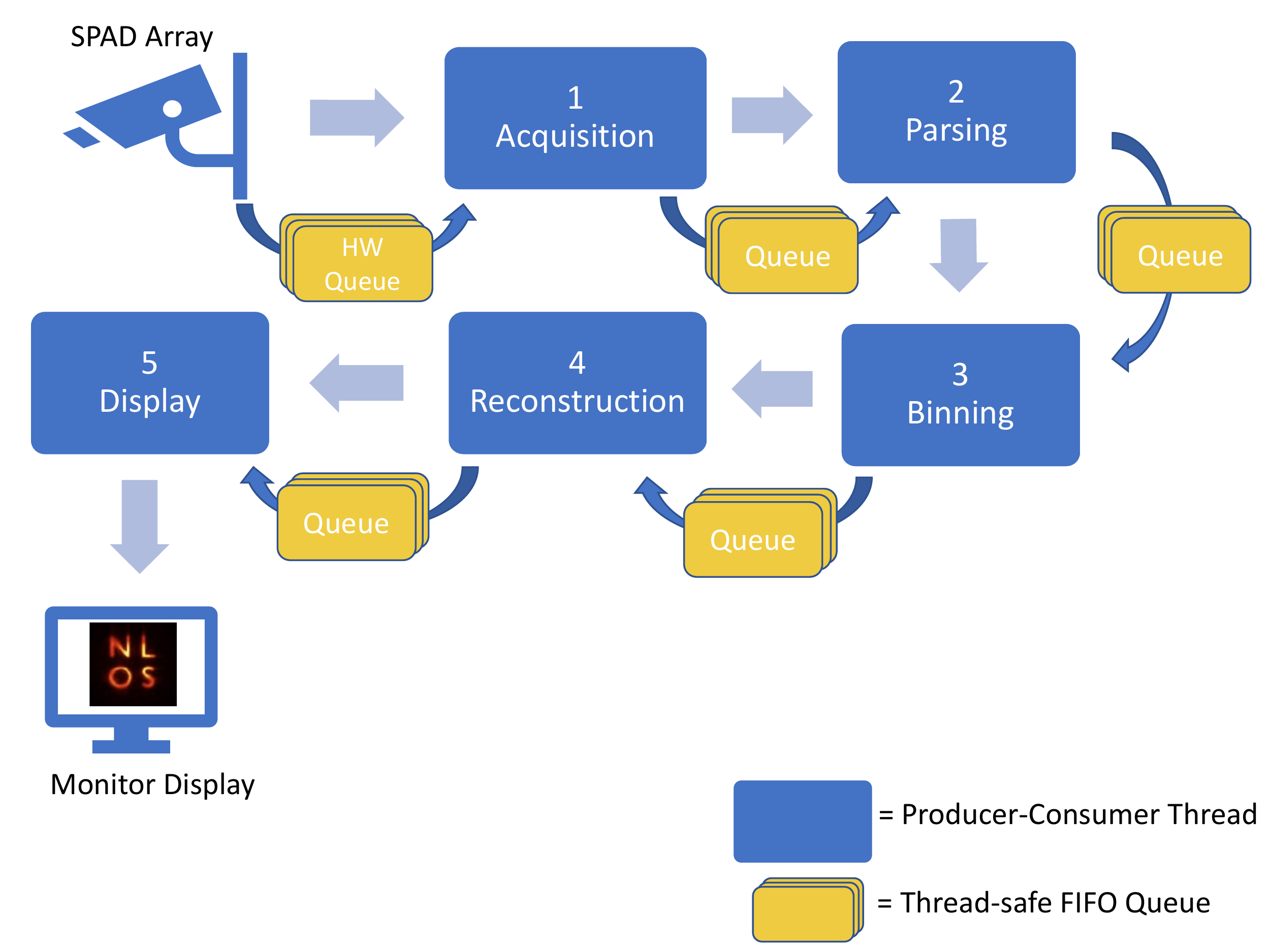}
    \caption{Software Block Diagram of Reconstruction Pipeline.}
    \label{fig:pipeline}
\end{figure}

\begin{figure}[htb]
\centering
\includegraphics[width=0.5\linewidth]{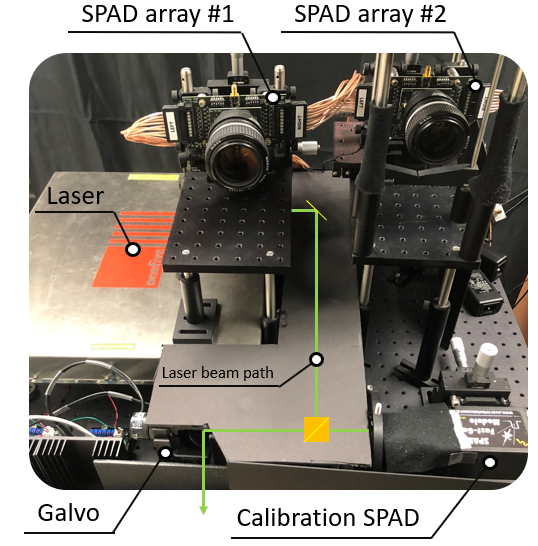}
\caption{\textbf{Real-time NLOS Hardware setup layout and laser beam path scheme.}}
\label{fig:hardware}
\end{figure}

\begin{figure}[htb]
\centering
\includegraphics[width=\linewidth]{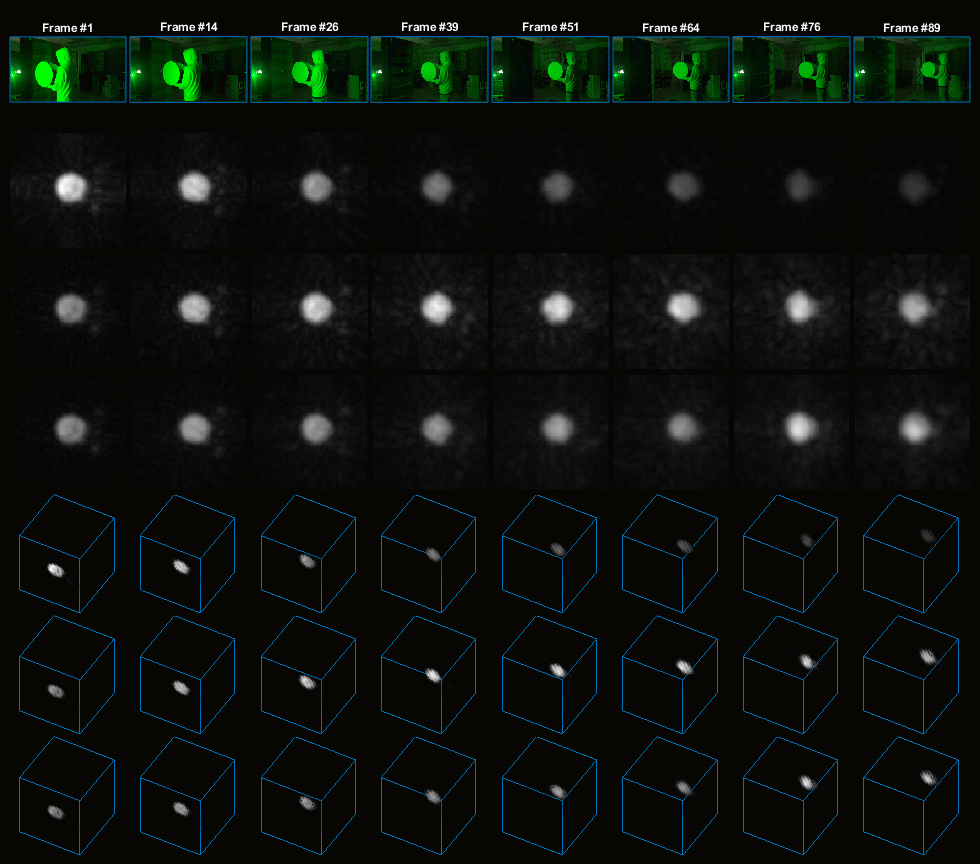}
\caption{\textbf{Real-time video: "Circle". } Ground truth, RSD, RSD with intensity correction, RSD with intensity correction and averaging.}
\label{fig:extra_1b}
\end{figure}

\begin{figure}[htb]
\centering
\includegraphics[width=\linewidth]{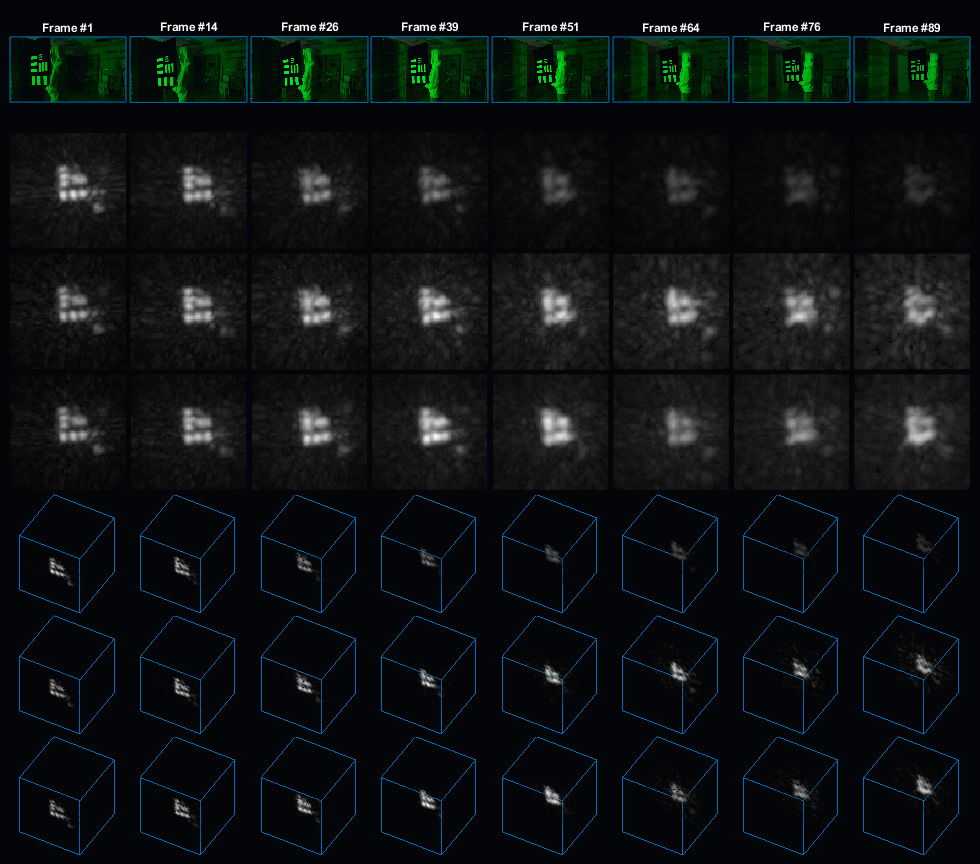}
\caption{\textbf{Real-time video: "Resolution bar". } Ground truth, RSD, RSD with intensity correction, RSD with intensity correction and averaging.}
\label{fig:extra_1c}
\end{figure}

\begin{figure}[htb]
\centering
\includegraphics[width=\linewidth]{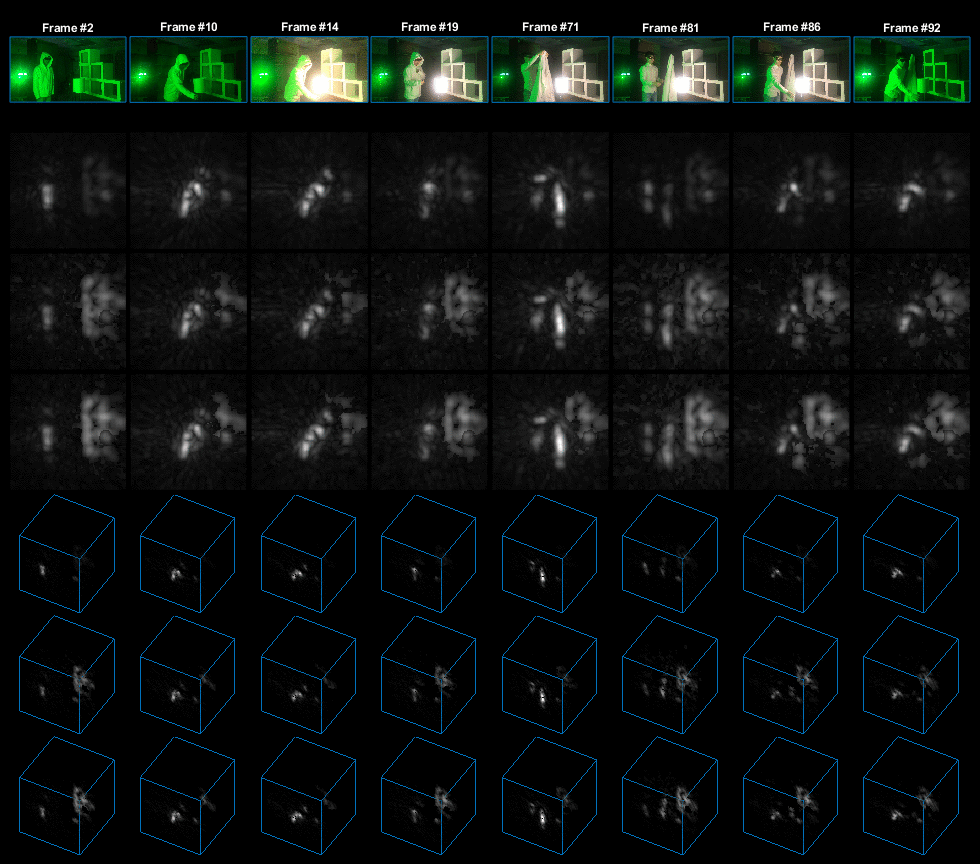}
\caption{\textbf{Real-time video: "At home". } Ground truth, RSD, RSD with intensity correction, RSD with intensity correction and averaging.}
\label{fig:extra_3}
\end{figure}

\begin{figure}[htb]
\centering
\includegraphics[width=\linewidth]{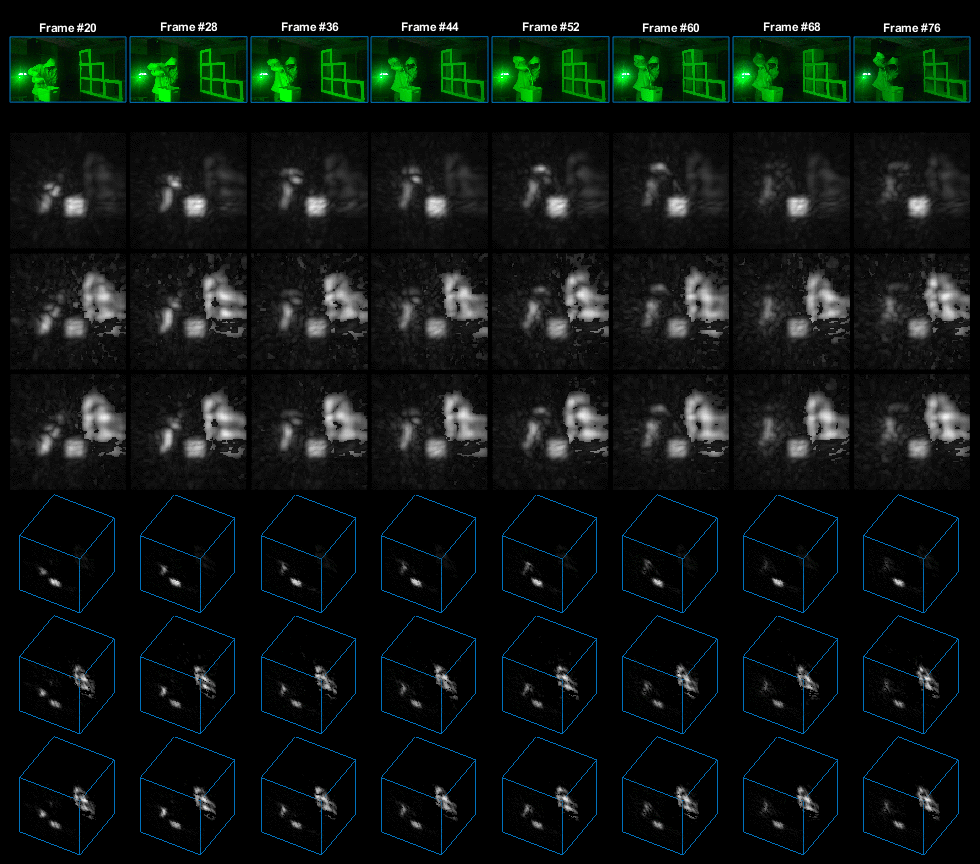}
\caption{\textbf{Real-time video: "Milk". } Ground truth, RSD, RSD with intensity correction, RSD with intensity correction and averaging.}
\label{fig:extra_4}
\end{figure}

\begin{figure}[htb]
\centering
\includegraphics[width=\linewidth]{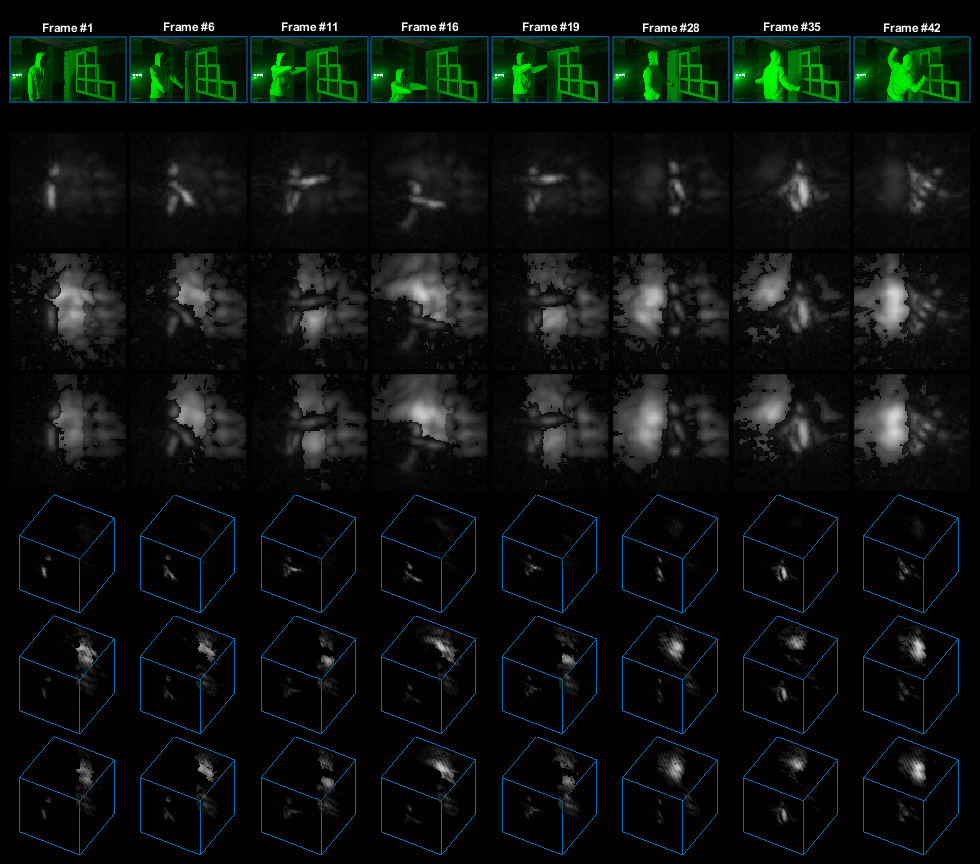}
\caption{\textbf{Real-time video: "Exercise". } Ground truth, RSD, RSD with intensity correction, RSD with intensity correction and averaging.}
\label{fig:extra_5}
\end{figure}

The software is designed using a multi-stage producer-consumer model with the processing broken up into five distinct stages. Each stage runs in a separate thread on the CPU, and is connected to its predecessor and successor stages by thread-safe FIFO queues. Data travels through the stages sequentially, with raw photon event records entering the first stage, and 2D images exiting the final stage. Each stage's thread runs in an infinite loop, performing the same sequence of tasks: wait for data to become available, retrieve the available data from its incoming queue, process the data, and finally submit the processed data to its outgoing queue. Figure \ref{fig:pipeline} shows a block diagram of the design. The five stages of processing are Acquisition, Parsing, Binning, Reconstruction, and Display. The staged pipeline and multithreaded model allows the entire pipeline to always remain full and working. That is, while frame $n$ is being displayed, simultaneously frame $n+1$ is being reconstructed, frame $n+2$ is being binned, frame $n+3$ is being parsed, and photon events for what will become frame $n+4$ are being collected. During properly tuned execution, the queues between stages never have more than a single entry waiting for processing, but serve primarily to decouple the processing of each stage. The implementation details of each stage are now described briefly.

\begin{description}
\item [Acquisition] The first stage directly connects to the HydraHarp API to retrieve raw photon timing records in the T3 format (4 bytes per photon record) via the vendor-provided USB3 driver. In each iteration of its infinite loop, the thread polls the hardware driver for all available photon records that have accumulated in a driver-side queue. Beginning-of-frame and end-of-frame markers are encoded inline with the photon events in the T3 record format. After retrieving all available records from the hardware FIFO, the array of raw (unparsed) T3 records is passed to the next stage by enqueuing the records into this stage's outgoing queue, and the thread repeats its process of polling the hardware again.

\item [Parsing] The parsing thread retrieves arrays of raw photon records from its incoming queue and unpacks each T3 record into usable photon information. This includes de-multiplexing the SPAD channels, calculating the grid-index of the photon based on galvo time, adjusting photon arrival timing based on physical SPAD geometry, and searching for start-of-frame and end-of-frame markers. Having found these markers, this stage packages the newly-calculated tuples of \texttt{(grid\_index, photon\_timing)} into an array representing a single discrete image frame and submits this array to its outgoing queue.

\item [Binning] The binning thread receives an entire frame of pre-processed tuples containing the grid indices and timings of each photon's arrival. These records are binned directly into a frequency domain histogram (FDH). See Algorithm \ref{alg:FDH} for details. To increase performance, we use OpenMP to enlist all available CPU threads to perform the frequency \texttt{for} loop in parallel, as there are no data write hazards on this loop. We achieve good cache-coherency due to the chosen FDH memory layout.  To further increase performance, we pre-multiply the frequency and omega values, and we discretize the photon arrival times to enable use of a pre-calculated lookup-table of $\sin$ and $\cos$ values that fits entirely within cache. The output of this stage is a FDH for a single frame.

\item [Reconstruction] The reconstruction thread dequeues a frame's FDH from its incoming queue and immediately transfers the FDH into GPU memory. Using an RSD kernel that has been pre-computed at application startup based on scene parameters, a sequence of CUDA kernels are executed to perform the Fast RSD algorithm's FFT, convolution with kernel, inverse FFT, and slice selection. The previous 3 reconstructed image cubes are held in memory and the resulting 2D image is formed by the depth dependent time averaging scheme described above. The resulting 2D image for this frame is then moved from GPU memory to main system memory and is enqueued in this stage's outgoing queue.

\item [Display] The final stages of the pipeline dequeues 2D images from its incoming queue. The 2D image is normalized, color-mapped, rotated, and scaled for display. The image is then displayed to the user, and the thread resumes waiting for the next 2D image to arrive in its incoming queue. 

\end{description}

\section*{References}
\bibliographystyle{unsrt}
\bibliography{ref}

\section*{Acknowledgements}
This work was funded by DARPA through the DARPA REVEAL project (HR0011-16-C-0025), and National Science Foundation grants NSF IIS-2008584, CCF-1812944, and IIS-1763638.

\section*{Author contributions statement}
A.V., J.H.N. and X.L. conceived the method, E.B., J.H.N. and E.S. implemented the real-time pipeline. S.B. developed the SNR models. J.H.N. built the experimental setup and developed the simulations. J.H.N., S.B. and A.V. analyzed the results. S.B. and X.L. helped with the system calibration. A.V. coordinated all aspects of the project. All authors contributed to the writing.  

\section*{Competing interests}
The authors declare no competing interests.

\newpage
 
\newpage

\newpage

\newpage

\newpage

\newpage

\newpage

\end{document}